\pdfoutput=1

\documentclass[11pt]{article}

\usepackage[final]{acl}

\usepackage{times}
\usepackage{latexsym}
\usepackage{lipsum}

\usepackage[T1]{fontenc}

\usepackage[utf8]{inputenc}

\usepackage{microtype}

\usepackage{inconsolata}

\usepackage{graphicx}

\usepackage{amssymb}
\usepackage{tabularray}
\usepackage{amsmath}
\usepackage{booktabs}
\usepackage{multirow}
\usepackage{array}
\usepackage{colortbl}
\usepackage{subcaption}
\usepackage[normalem]{ulem}
\usepackage{CJKutf8}
\usepackage{algorithm}
\usepackage{algpseudocode}
\usepackage{tcolorbox}
\usepackage{pgf} 
\usepackage{inconsolata}
\usepackage{appendix}
\usepackage{color}
\usepackage{longtable}
\usepackage{ragged2e}
\usepackage{rotating}
\usepackage{caption}
\usepackage{float}                  
\usepackage{overpic}                
\usepackage{tablefootnote}
\usepackage{wrapfig}
\usepackage{enumitem}
\usepackage{tcolorbox}

\title{
On-Policy Self-Alignment with Fine-grained Knowledge Feedback \\ for Hallucination Mitigation
}

\author{
  Xueru Wen${}^{1,2}$,
  Jie Lou${}^{3}$,
  Xinyu Lu${}^{1,2}$,
  Ji Yuqiu${}^{3}$,
  Xinyan Guan${}^{1,2}$,
  Yaojie Lu${}^{1*}$,
  \\
  {\bf Hongyu Lin${}^{1}$,}
  {\bf Ben He${}^{1,2*}$,}
  {\bf Xianpei Han${}^{1,2}$,}
  {\bf Debing Zhang${}^{3}$,}
  {\bf Le Sun${}^{1,2}$}
  \\
  ${}^{1}$Chinese Information Processing Laboratory, \\
  Institute of Software, Chinese Academy of Sciences, Beijing, China \\
  ${}^{2}$University of Chinese Academy of Sciences, Beijing, China \\
  ${}^{3}$Xiaohongshu Inc \\
  \tt {\{wenxueru2022,luxinyu2021,guanxinyan2022\}@iscas.ac.cn} \\
  \tt {\{luyaojie,hongyu,sunle,xianpei\}@iscas.ac.cn} \\
  \tt{benhe@ucas.edu.cn loujie0822@gmail.com dengyang@xiaohongshu.com}
}

\begin{document}
\maketitle
\renewcommand{\thefootnote}{}
\footnotetext{\textsuperscript{*}Corresponding authors.}
\renewcommand{\thefootnote}{\arabic{footnote}}

\begin{abstract}
Hallucination occurs when large language models exhibit behavior that deviates from the boundaries of their knowledge during response generation.
To address this critical issue, previous learning-based methods attempt to finetune models but are limited by off-policy sampling and coarse-grained feedback.
In this paper, we present \textit{\b{R}einforcement \b{L}earning \b{f}or \b{H}allucination} (RLFH), an on-policy self-alignment approach that enables LLMs to actively explore their knowledge boundaries and self-correct generation behavior through fine-grained feedback signals.
RLFH introduces a self-assessment framework where the policy serves as its own judge. 
Through this framework, responses are automatically decomposed into atomic facts and their truthfulness and informativeness are assessed against external knowledge sources.
The resulting fine-grained feedback at the statement level are then converted into token-level dense reward signals. 
This enables online reinforcement learning to achieve precise and timely optimization without human intervention.
Comprehensive evaluations on HotpotQA, SQuADv2, and Biography benchmarks validate RLFH's effectiveness in hallucination mitigation.
\end{abstract}

\section{Introduction}
\begin{figure*}[t] 
\centering 
\includegraphics[width=\textwidth]{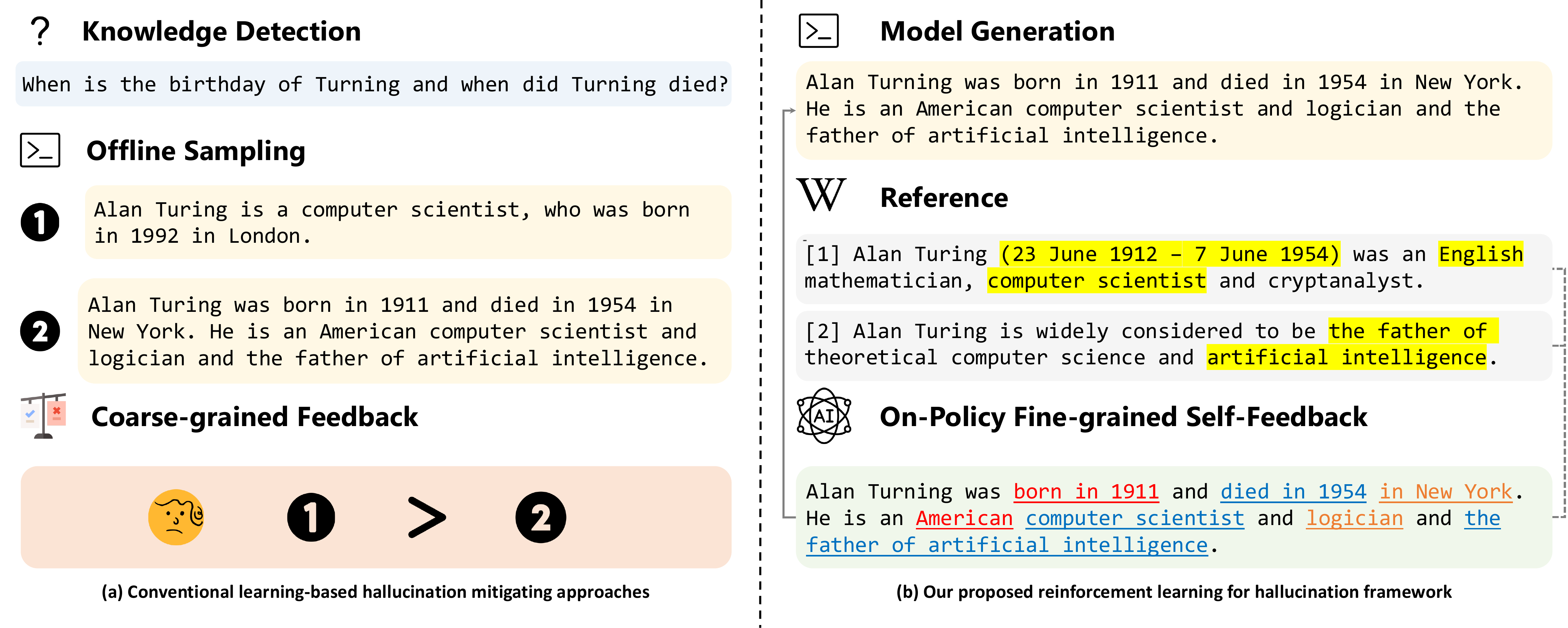} 
\caption{
The figure illustrates the hallucinatory case and several hallucination mitigation methodologies. 
The factual information within the text is underlined.
False content is highlighted in \textcolor{red}{red}, whereas accurate facts are indicated in \textcolor{blue}{blue}.
Statements with uncertain veracity are marked in \textcolor{orange}{orange}.
} 
\label{fig:demo} 
\end{figure*}

Large language models (LLMs) have demonstrated capabilities in generating fluent and plausible responses.
However, these models occasionally fabricate facts in their responses, referred to as \textbf{\textit{hallucination}}.
The crux of hallucination is \textit{the misalignment between models' generation and their internal knowledge} \citep{xu2024rejection}.
This misalignment manifests in various ways. For instance, as shown in Figure \ref{fig:demo}, the response of LLMs about "Turing" contains erroneous factual information, such as stating that he was born in 1911 and was American.
More broadly, these hallucinations can be categorized into several types: (1) \textbf{misleading responses}, when the model inaccurately answers questions within its knowledge boundary; (2) \textbf{reckless attempts}, when the model responds to queries beyond its knowledge; and (3) \textbf{evasive ignorance}, when the model refrains from providing answers despite possessing the knowledge.
Unfortunately, due to the opaque nature of model knowledge, we can only observe erroneous model responses or their refusal to respond, without accurately determining whether they have experienced hallucinations.

Recent studies have attempted to mitigate hallucination in large language models via learning-based and editing-based approaches.
Learning-based methods first detect the model's knowledge boundaries and then finetune it with carefully curated feedback data.
However, these methods face several challenges. 
First, due to off-policy data sampling \citep{zhang2024selfalignment,wan2024knowledge,lin2024flame}, they experience distribution shifts, resulting in suboptimal models \citep{tang2024understanding}.
Second, coarse-grained instance-level feedback \citep{sun2022contrastive,tian2023finetuning,kang2024unfamiliar} fails to precisely pinpoint the hallucinations, as a single response may contain both correct and incorrect facts.
Finally, given our limited understanding of how models learn and express knowledge, existing knowledge detection techniques \citep{zhang2023rtuning,cheng2024ai,yang2023alignment} may produce inconsistent results, thus failing to accurately reflect the model's knowledge boundaries.
In contrast, editing-based methods \citep{gou2023critic, manakul-etal-2023-selfcheckgpt} first generate content and then edit it based on external knowledge.
These methods face two fundamental limitations: they rely heavily on external knowledge sources which are inherently limited in scope, and more importantly, they only correct output content without addressing the underlying issue of how models utilize their internal knowledge.
In general, hallucination mitigation requires fine-grained feedback tailored to the online model, which enables the model to effectively explore its knowledge boundaries and form reliable behavior.

In this paper, we present \textit{\b{R}einforcement \b{L}earning \b{f}or \b{H}allucination} (RLFH), an on-policy self-alignment approach that uses fine-grained feedback for hallucination mitigation.
Our approach enables LLMs themselves to explore their own knowledge boundaries through fine-grained, on-policy feedback.
With this direct feedback about their internal knowledge state, LLMs learn to balance knowledge usage and thus reduce hallucination.
Specifically, RLFH guides LLMs to first generate initial responses and then conduct a self-verification process.
The responses are decomposed into atomic facts and then undergo self-assessment against external knowledge sources.
During assessment, the current model determines whether an atomic fact aligns with the facts described in the ground-truth document and assesses the informativeness of the fact.
The resulting statement-level assessments are converted into token-level dense reward signals. 
These precise, real-time rewards enable RLFH to directly optimize on-policy behavior through online reinforcement learning.
By having the policy serve as its own judge, we construct a self-driven fact assessment framework that enables timely and low-cost reward signal collection for on-policy optimization without human intervention.

The main contributions are as follows:
\begin{enumerate}
    \item[\textbf{1)}] We propose RLFH, an on-policy self-alignment framework that enables LLMs to actively explore their own knowledge boundaries and self-correct generation behavior through fine-grained feedback signals.
    \item[\textbf{2)}] We design a self-assessment framework where the policy serves as its own judge, automatically decomposing responses into atomic facts and evaluating their truthfulness and informativeness. This framework generates fine-grained knowledge feedback in real-time and provides token-level dense reward signals for online reinforcement learning.
    \item[\textbf{3)}] Comprehensive evaluations on HotpotQA, SQuADv2, and Biography present significant improvements of RLFH over both base models and existing hallucination mitigation approaches, showing the method's effectiveness\footnote{The code and data associated with this work are available at \url{https://github.com/AlignRM/RLFH}.}.
\end{enumerate}

\section{Related Works}
\subsection{Hallucination Mitigation}
Prior research \citep{zhang2023sirens,ye2023cognitive,tonmoy2024comprehensive} has been dedicated to addressing the hallucination of LLMs.
Some studies focus on reducing errors \citep{wang-2019-revisiting, parikh-etal-2020-totto} and supplementing missing knowledge \citep{ji2023domainspecific} during data curation.
Other works mitigate hallucination in either pre- or post-generation by retrieving external knowledge \citep{peng2023check, li2023self, gou2023critic} or exploiting self-consistency \citep{manakul-etal-2023-selfcheckgpt, shi2023trusting, lee2023factuality}.
Recent studies focus on investigating the essence of the hallucination \citep{yu2024mechanisms,jiang2024large} and resort to improving the model's factuality during the alignment stage.
These works focus on resolving the inconsistency between the model's generation and its internalized knowledge \citep{xu2024rejection} through knowledge detection and coarse-grained feedback.
Typically, these works attempt to delineate the boundary of model knowledge through explicit prompting \citep{zhang2023rtuning,yang2023alignment,cheng2024ai,wan2024knowledge}, self eliciting \citep{chen2024grath,lin2024flame}, self-evaluation \citep{zhang2024selfalignment} or probing the model's internal states \citep{liang2024learning}.
Based on such knowledge boundary detection, the data is meticulously crafted to align with the model's knowledge scope. 
Subsequently, the model is fine-tuned with coarse-grained feedback, which inspects the truthfulness of the response as a whole \citep{sun2022contrastive,tian2023finetuning,kang2024unfamiliar,huang2024factalignlongformfactualityalignment,gao2024honestllmhonesthelpfullarge}.

\subsection{Reinforcement Learning from Human Feedback}
Reinforcement Learning from Human Feedback \citep{NEURIPS2020_1f89885d, ouyang2022training} has emerged as a noteworthy approach for LLM alignment.
Given the instability of reinforcement learning, some research \citep{lu_quark_2022, rafailov_direct_2023, dong_raft_2023} has attempted to learn preferences directly from labeled data.
In addition to sparse rewards, some works have explored designing more instructive rewards.
One line of works \citep{wu_fine-grained_2023, lightman2023lets, chen2024improving, cao2024drlc} is dedicated to the acquisition of dense rewards.
Another line of works \citep{ramé2023rewarded, eisenstein2023helping, coste2024reward, ramé2024warm} concentrates on ensemble multiple reward models.
Finally, few studies \citep{wu_fine-grained_2023, tian2023finetuning, liang2024learning} have explicitly targeted truthfulness.

\section{Reinforcement Learning for Hallucination}
\label{sef:method}
\begin{figure*}[t] 
\centering 
\includegraphics[width=\textwidth]{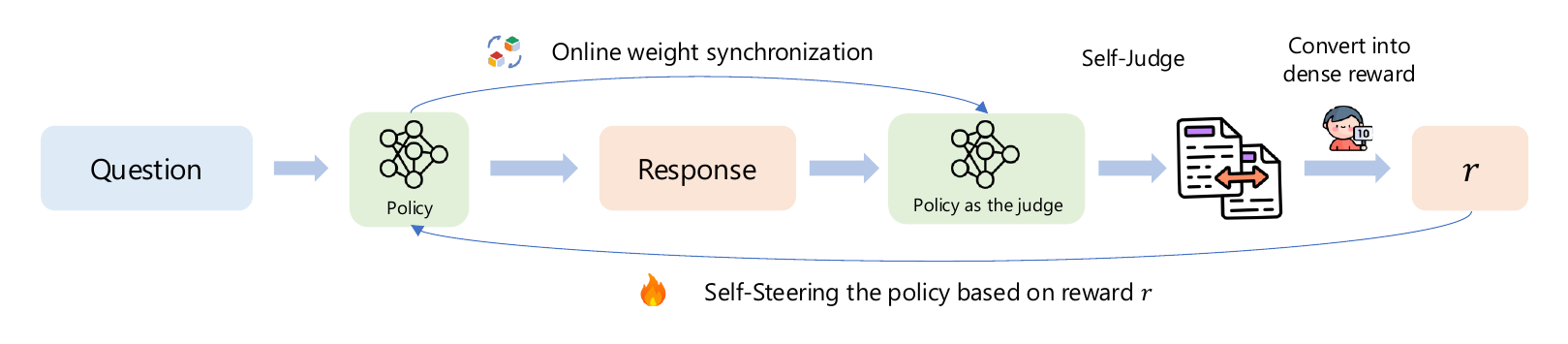} 
\caption{
A diagram illustrating the steps of our algorithm: (1) Sampling response from tuning model, (2) Policy acting as a judge model performing self-assessment to collect fine-grained knowledge feedback, and (3) Converting the language-form feedback into token-level dense reward for reinforcement learning.
} 
\label{fig:algo} 
\end{figure*}

Given the train prompt set $\mathcal{X} = \{x_{1}, x_{2}, ..., x_{|\mathcal{X}|}\}$, the policy model $\pi$ being optimized, and the reference document set $\mathcal{D} = \{d_{1}, d_{2}, ..., d_{|\mathcal{D}|}\}$, this section demonstrates the procedure of our approach, shown in Figure \ref{fig:algo}.
Here's a detailed breakdown of each step:
1) \textbf{Response Generation}: Given the prompt $x_{i}$, the policy model $\pi$ generates a corresponding response $y_{i}$. This step involves the model using its current policy to produce an output based on the input prompt.
2) \textbf{Fine-grained Feedback from Policy as Judge} ($\S$\ref{sec:feedback}): The policy $\pi$, acting as its own judge, evaluates the generated response $y_{i}$ through atomic fact decomposition and verification against the reference document set $\mathcal{D}$, providing fine-grained feedback $\mathcal{E}$ at the statement level.
3) \textbf{On-Policy Optimization with Token-level Reward} ($\S$\ref{sec:ppo_update}):  The detailed feedback $\mathcal{E}$ is translated into token-level rewards $r$. 
These rewards are then used to update the policy model $\pi$ using online reinforcement learning algorithm, ensuring that the model learns to reduce hallucinations effectively.

\subsection{Fine-grained Feedback from Policy as the Judge}
\label{sec:feedback}
Given the prompt $x_i$ and its corresponding response $y_i$, RLFH enables the policy $\mathcal{\pi}$ to conduct self-assessment, providing fine-grained feedback on truthfulness and informativeness at the statement level.
Specifically, the policy $\pi$ first decomposes the response $y_i$ into atomic statements $\mathcal{E}_i = \left\{e_1,e_2,...,e_{|\mathcal{E}_i|}\right\}$, where each statement $e_j$ represents an atomic fact in the response.
Subsequently, acting as its own judge, the policy verifies each atomic statement $e_{i}$ against the reference document to provide fine-grained feedback.

\subsubsection{Statement Extraction} 
\label{sec:extra}
Given a query $x$ and its corresponding output $y$, we leverage the current policy model $\pi$ to partition responses and extract atomic factual statements in a hierarchical manner.
Specifically, $\pi$ initially divides the response into sentences $\{s_i\}_{i=1}^{M}$ and then extracts all valid factual statements $\{e_{ij}\}_{j=1}^{N_i}$ from each sentence $s_i$.
There are two reasons for this hierarchical approach: 
(1) Splitting the response into sentences before extracting statements consistently yields finer granularity;
(2) Extracting statements sentence-by-sentence facilitates the conversion from language-form annotation to token-level dense reward.
After performing extraction, we further filter out sentences without valid statements to mitigate potential noise. 

\subsubsection{Factual Verification}
The policy model $\pi$ evaluates the truthfulness of the extracted factual statements by comparing them with external knowledge sources.
For each statement $e$, we retrieve relevant supporting contexts $\{c_i\}_{i=1}^{L} \subset \mathcal{D}$ from the reference document set $\mathcal{D}$.
With these supporting contexts, the policy model $\pi$ conducts statement verification as a reading comprehension task, represented as:
\begin{equation}
    k_\text{truth} = \mathcal{\pi}(e,\{c_{i}\}_{i=1}^{L})
\end{equation}
Specifically, the policy model $\pi$ classifies each statement into the following labels:
1) \textit{Correct}: correct statement supported by evidence;
2) \textit{Hedged Correct}: accurate statement with uncertainty;
3) \textit{Vague}: statement with uncertain truthfulness;
4) \textit{Hedged Wrong}: false statement with uncertainty;
5) \textit{Wrong}: statement contradicted by evidence.
We introduce the "\textit{Vague}" category to handle statements whose truthfulness cannot be verified based on reference documents due to limited supporting materials or unclear evidence.

\subsubsection{Informativeness Assessment}
In addition to correctness, the policy model $\pi$ further evaluates the informativeness of the statements.
We assess each statement's informativeness on a five-point scale, ranging from providing crucial information (+5) to containing minimal relevant details (+1).
Unlike the individual statement verification process, assessing informativeness requires considering the original query $x$ and response $y$.
This is because informativeness evaluation requires considering the overall context and content comprehensiveness, rather than just individual statements' truthfulness.
This process can be denoted as:
\begin{equation}
    k^i_{\text{info}}=\mathcal{\pi}\left(x,y,e_i\right)    
\end{equation}
The introduction of informativeness prevents the trivial hack that the model either rejects the majority of responses or produces only brief answers, both of which are undesirable outcomes.

\subsection{On-Policy Optimization with Token-level Reward}
\label{sec:ppo_update}
Given the fine-grained, statement-level feedback from the policy-as-judge framework, we trace the atomic facts' assessment back to the original response $y$ and construct token-level dense reward signals $r$ for direct optimization.
Finally, we adopt online reinforcement algorithm with these token-level reward signals to mitigate hallucination in the model's generation behavior.

\begin{figure*}[t] 
\centering 
\includegraphics[width=\textwidth]{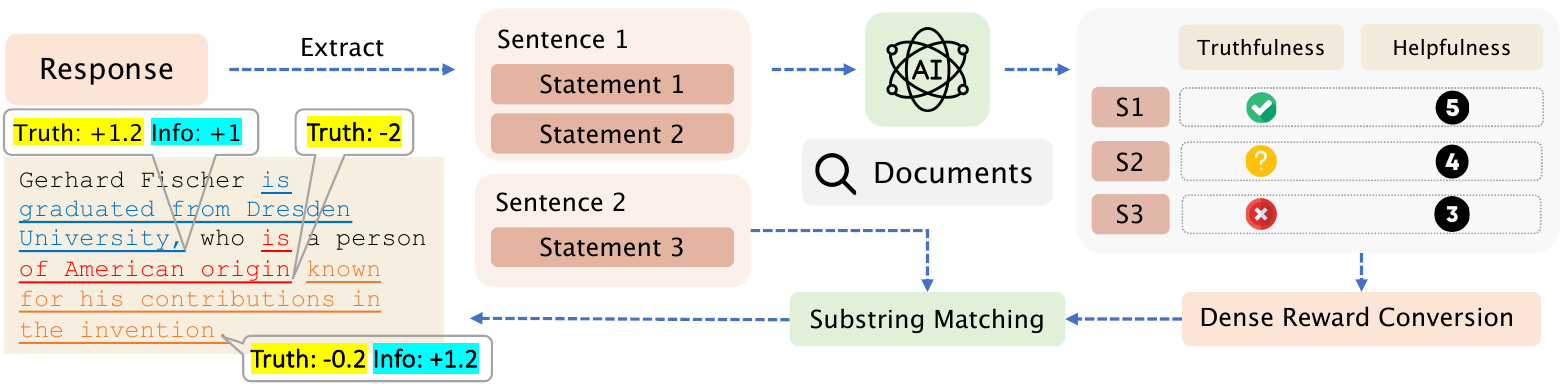} 
\caption{
A schematic representation of fine-grained feedback and token-level reward strategy methodology is presented. Initially, the statements are extracted in a hierarchical fashion. Subsequently, the veracity and utility of each statement are assessed. Ultimately, the structured feedback is mapped back into a dense reward via the Longest Common Subsequence (LCS) algorithm.
} 
\label{fig:reward} 
\end{figure*}

\subsubsection{Dense Reward Conversion}
We represent the informativeness and truthfulness of the response $y$ through the dense reward conversion presented in Figure \ref{fig:reward}.
Due to the mutually exclusive nature \citep{xu2024rejection} of these two objectives, the model should learn a appropriate strategy for utilizing its internal knowledge to balance the pursuits of truthfulness and informativeness.

\paragraph{Truthfulness}
For each extracted statement, we assign a truthfulness reward computed as follows:
\begin{equation}
    r_{\text{truth}} = \alpha f(k_{\text{truth}})|g(k_{\text{info}})|
\end{equation}
where $f$ and $g$ represent manually designed functions that transform the labels $k$ into scalar values.
In principle, $f$ gives a positive reward to the correct statement and a negative reward to the unverifiable or false statement.
Due to the hallucination snowball effect \citep{zhang2023language}, where critical errors can lead to magnified hallucinations, $g$ is included to diversify the importance of different statements.
The absolute value function applied to $g$ preserves the sign of function $f$'s output.
The coefficient $\alpha$ balances between the truthfulness reward and the helpfulness reward.

The reward $r_{\text{truth}}$ is mapped back to the token sequence of the response $y$ using the hierarchical structure constructed in the aforementioned annotations.
Specifically, we first use the Longest Common Subsequence algorithm to locate each statement $e_{ij}$ within its originating sentence $s_i$.
Subsequently, each sentence $s_i$ is mapped back to the model's response $y$ through the Longest Common Substring algorithm.
Finally, the reward $r_{\text{truth}}$ is assigned to the token in the response corresponding to the statement's last character position.

\paragraph{Informativeness}
For each sentence, we assign an informativeness reward based on the statements it encompasses as follows:
\begin{equation}
    r_{\text{info}} = \beta \log \left(\mu + {\text{max}\left(\epsilon, \sum_i^N g\left(k^i_{\text{info}}\right)\right)}\right)
\end{equation}
In this equation, $N$ denotes the total number of statements within a sentence, while $\epsilon$ and $\mu$ form the minimum reward threshold serving to penalize non-informative statements. 
As indicated by the equation, the reward increases with the number of statements in a sentence and their respective informativeness. 
However, the rate of growth of the reward decreases rapidly. 
Conversely, the penalty for producing non-informative statements escalates swiftly.
We apply the same mapping method as used for the truthfulness reward to trace the reward value back to the response token sequence.

\subsubsection{Online Reinforcement Learning} 
Given the reward function, the policy model $\pi$ is optimized through online reinforcement learning to maximize the reward expectation:
\begin{equation}
    \arg \max \limits_{\pi}\mathbb{E}_{x\sim\mathcal{X},y\sim\pi}\left[\sum_{i=1}^T r\left(y_t,\left(x,y_{1:t}\right)\right)\right]
\end{equation}
Specifically, we first sample the prompt $x$ and corresponding response $y$.
Subsequently, the policy model $\pi$ itself serves as the judge to provide fine-grained feedback through the assessment framework.
This feedback is then converted into token-level dense reward $r=\left[r_1,r_2,...,r_T\right]$, where $T$ denotes the total length of the response $y$.
Finally, we employ this reward $r$ to update the policy model $\pi$ by the Proximal Policy Optimization \citep{DBLP:journals/corr/SchulmanWDRK17} algorithm.

\section{Experiment}
\begin{table*}[ht]
\centering
\resizebox{\linewidth}{!}{%
\begin{tabular}{lccccccccccccccc} 
\toprule
\multicolumn{1}{c}{\multirow{2}{*}{\textbf{Model}}} & \textbf{\textbf{Avg.}} & \multicolumn{4}{c}{\textbf{HotpotQA}} &  & \multicolumn{4}{c}{\textbf{SQuADv2}} &  & \multicolumn{4}{c}{\textbf{Biography}} \\ 
\cmidrule{3-6}\cmidrule{8-11}\cmidrule{13-16}
\multicolumn{1}{c}{} & \textbf{\textbf{\textbf{\textbf{Score}}}} & \textbf{\#Cor.} & \textbf{\#Inc.} & \textbf{\%Res.} & \textbf{Score} &  & \textbf{\#Cor.} & \textbf{\#Inc.} & \textbf{\%Res.} & \textbf{Score} &  & \textbf{\#Cor.} & \textbf{\#Inc.} & \textbf{\%Res.} & \textbf{Score} \\ 
\hline
\multicolumn{16}{c}{\textit{Open-source Models}} \\ 
\midrule
DeepSeekV2-Lite & 0.618 & 15.4 & 9.22 & 0.96 & 0.642 &  & 23.6 & 7.09 & 0.98 & 0.754 &  & 28.0 & 32.4 & 0.96 & 0.458 \\
Falcon3-10B & 0.593 & 5.14 & 3.06 & 0.90 & 0.608 &  & 11.1 & 2.18 & 0.96 & 0.813 &  & 13.5 & 22.6 & 1.00 & 0.357 \\
Ministral-8B & 0.591 & 7.36 & 3.81 & 0.96 & 0.633 &  & 15.7 & 4.26 & 0.82 & 0.761 &  & 22.7 & 37.3 & 1.00 & 0.378 \\
Yi-1.5-9B & 0.536 & 12.7 & 12.0 & 1.00 & 0.533 &  & 28.9 & 10.0 & 1.00 & 0.734 &  & 29.4 & 56.9 & 1.00 & 0.340 \\ 
Qwen2.5-7B & 0.638 & 9.13 & 4.80 & 0.93 & 0.634 &  & 21.1 & 4.82 & 0.97 & 0.813 &  & 20.9 & 23.1 & 0.73 & 0.467 \\
Llama-3.1-8B & 0.639 & 4.57 & 2.44 & 0.99 & 0.652 &  & 22.8 & 6.02 & 0.98 & 0.777 &  & 17.6 & 12.5 & 0.84 & 0.487 \\
\midrule
\multicolumn{16}{c}{\textit{Baselines based on Llama3.1-8B-Instruct}} \\ 
\midrule
DOLA & 0.546 & 6.61 & 6.00 & 0.90 & 0.524 &  & 22.6 & 8.61 & 0.97 & 0.713 &  & 15.6 & 20.4 & 0.84 & 0.399 \\
ITI & 0.646 & 4.48 & 1.91 & 0.99 & 0.649 &  & 19.2 & 5.03 & 0.98 & 0.776 &  & 19.1 & 16.1 & 0.90 & 0.512 \\
FACT$_\text{DPO}$ & 0.645 & 4.90 & 2.18 & 0.99 & 0.652 &  & 22.6 & 6.31 & 0.99 & 0.778 &  & 18.1 & 12.3 & 0.85 & 0.506 \\
FACT$_\text{SFT}$ & 0.653 & 2.49 & 1.31 & 1.00 & 0.635 &  & 17.2 & 4.60 & 1.00 & 0.783 &  & 5.7 & 4.0 & 0.99 & 0.541 \\ 
\midrule
\multicolumn{16}{c}{\textit{RLFH on Different Models}} \\ 
\midrule
RLFH$_\text{Qwen2.5-7B}$ & 0.668 & 7.30 & 3.66 & 0.90 & 0.651 &  & 17.3 & 3.55 & 0.96 & \textbf{0.830} &  & 17.5 & 15.5 & 0.59 & 0.523 \\
RLFH$_\text{Llama3.1-8B}$ & \textbf{0.686} & 6.23 & 2.10 & 1.00 & \textbf{0.714} &  & 21.2 & 5.32 & 1.00 & 0.786 &  & 17.3 & 11.0 & 0.79 & \textbf{0.558} \\
\bottomrule
\end{tabular}
}
\caption{Experiment results on HotpotQA, SQuADv2, and Biography.}
\label{tab:main_result}
\end{table*}

\subsection{Settings} \label{sec:exp_settings}
\paragraph{Datasets}
We employ three distinct datasets for our experiments.
Following the approach in \cite{min2023factscore}, we filter out prompts lacking corresponding wiki pages for both training and evaluation.
Additionally, we sample 10,000 questions from \textbf{HotpotQA} \citep{yang2018hotpotqa} and use the English Wikipedia from 04/01/2023 as the retrieval corpus for training.
We filter questions in \textbf{HotpotQA} with less than 5 words and sample 256 questions for evaluation.
We deduplicate questions in \textbf{SQuADv2} \citep{rajpurkar2016squad} based on their reference wiki pages, retaining 191 questions for out-of-distribution QA evaluation.
\textbf{Biography} dataset is identical to the one used in FactScore \citep{min2023factscore} for evaluating out-of-distribution performance on different forms of text.

\paragraph{Baselines}
We compare RLFH with two different types of baselines:
1) \textit{hallucination mitigation methods} based on the same initialized model Llama3.1-8B-Instruct, including decoding by contrasting layers (\textbf{DOLA}) \citep{chuang2023dola}, inference-time intervention (\textbf{ITI}) \citep{li2023inferencetime}, and factuality finetuning (\textbf{FACT}) \citep{tian2023finetuning} based on DPO \citep{rafailov2024directpreferenceoptimizationlanguage} and SFT;
2) \textit{advanced aligned models} of comparable size, including \textbf{Llama3.1-8B-Instruct} \citep{grattafiori2024llama3herdmodels}, \textbf{Qwen2.5B-7B-Instruct} \citep{qwen2025qwen25technicalreport}, \textbf{DeepSeekV2-Lite-Chat} \citep{deepseekv2}, \textbf{Falcon3-10B-Instruct} \citep{Falcon3}, and \textbf{Yi-1.5-9B-Chat} \citep{ai2025yiopenfoundationmodels}.

\paragraph{Evaluation}
We employ the FactScore \citep{min2023factscore} pipeline implemented with Qwen2.5-72B-Instruct\footnote{Discussion about the metric is provided in Appendix \ref{sec:metric_valid}.} \citep{qwen2025qwen25technicalreport} to evaluate the truthfulness and helpfulness of each generated response.
Following previous works \citep{tian2023finetuning,lin2024flame}, we adopted FactScore without length penalty, which represents the average accuracy of statements in the response.
For each dataset, we report the number of correct and relevant facts (\#Cor.), the number of inaccurate facts (\#Inc.), the ratio of responded questions (\%Res.), and the computed FactScore metrics (Score).

\begin{figure}[ht]
    \centering
    \includegraphics[width=\linewidth]{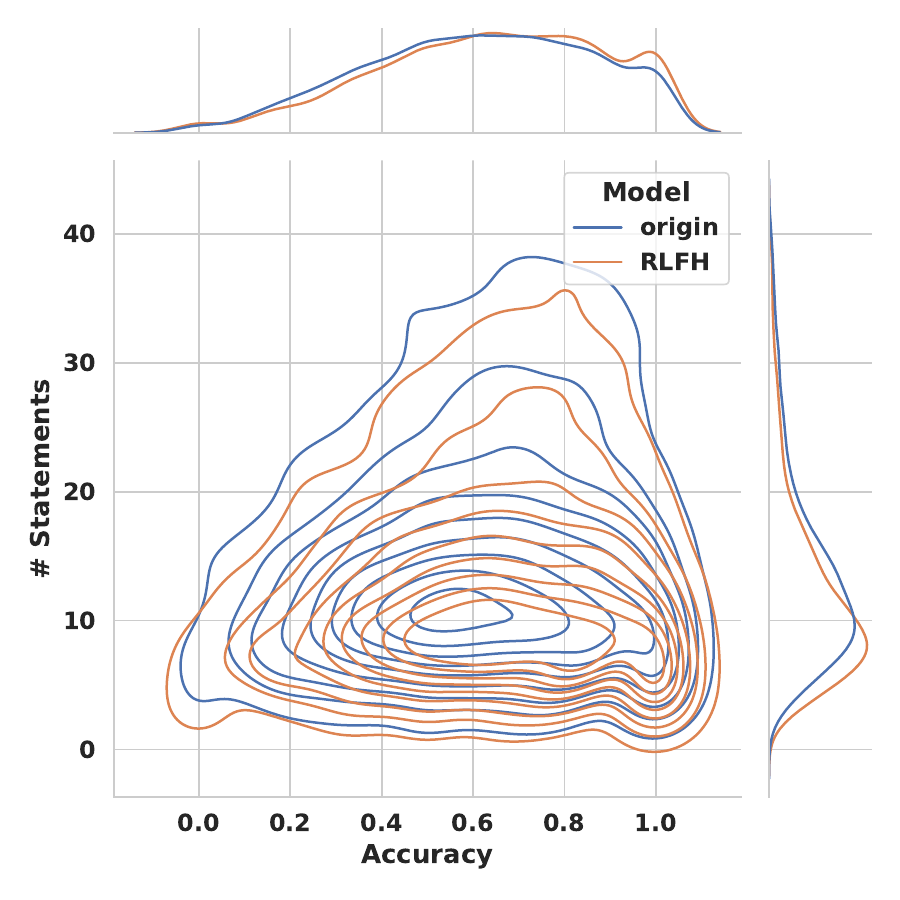}
    \caption{Distribution of statement accuracy versus count per response for Qwen2.5-7B-Instruct, comparing the base model and RLFH-tuned model.}
    \label{fig:join}
\end{figure}

\paragraph{Implementation}
Our training implementations are developed based on OpenRLHF \citep{hu2024openrlhfeasytousescalablehighperformance}.
The base model utilized including Qwen2.5B-7B-Instruct and Llama3.1-8B-Instruct. Detailed prompts for preforming the annotation pipeline are shown in Appendix \ref{sec:prompt}.
The hyperparameter settings are provided in Appendix \ref{sec:hyper}.

\subsection{Main Results}
\label{sec:main}
Table \ref{tab:main_result} presents the performance comparison between RLFH and all baselines on three datasets based on FactScore evaluation pipeline.

\begin{figure*}[t]
\centering
\subfloat[Distribution of correct statements.]{
    \includegraphics[width=0.33\textwidth]{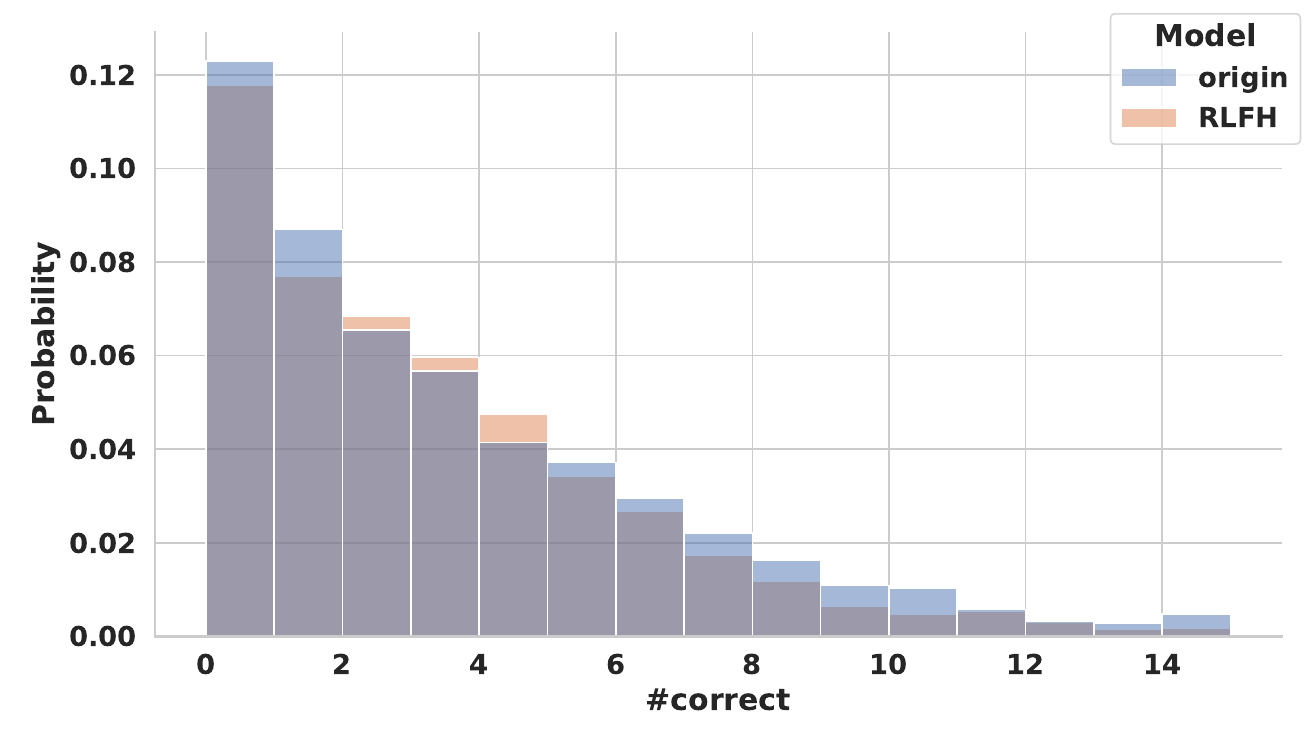}
    \label{fig:cor}
}
\subfloat[Distribution of vague statements.]{
    \includegraphics[width=0.33\textwidth]{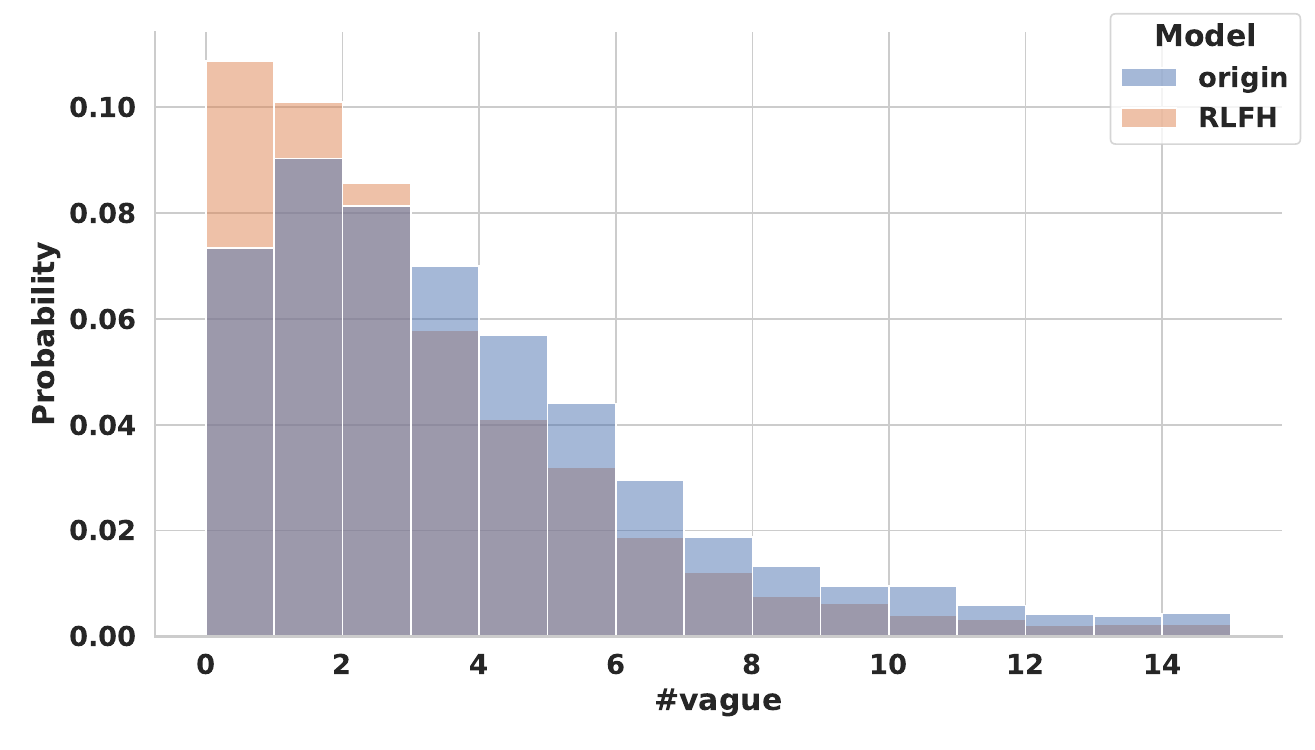}
    \label{fig:vague}
}
\subfloat[Distribution of wrong statements.]{
    \includegraphics[width=0.33\textwidth]{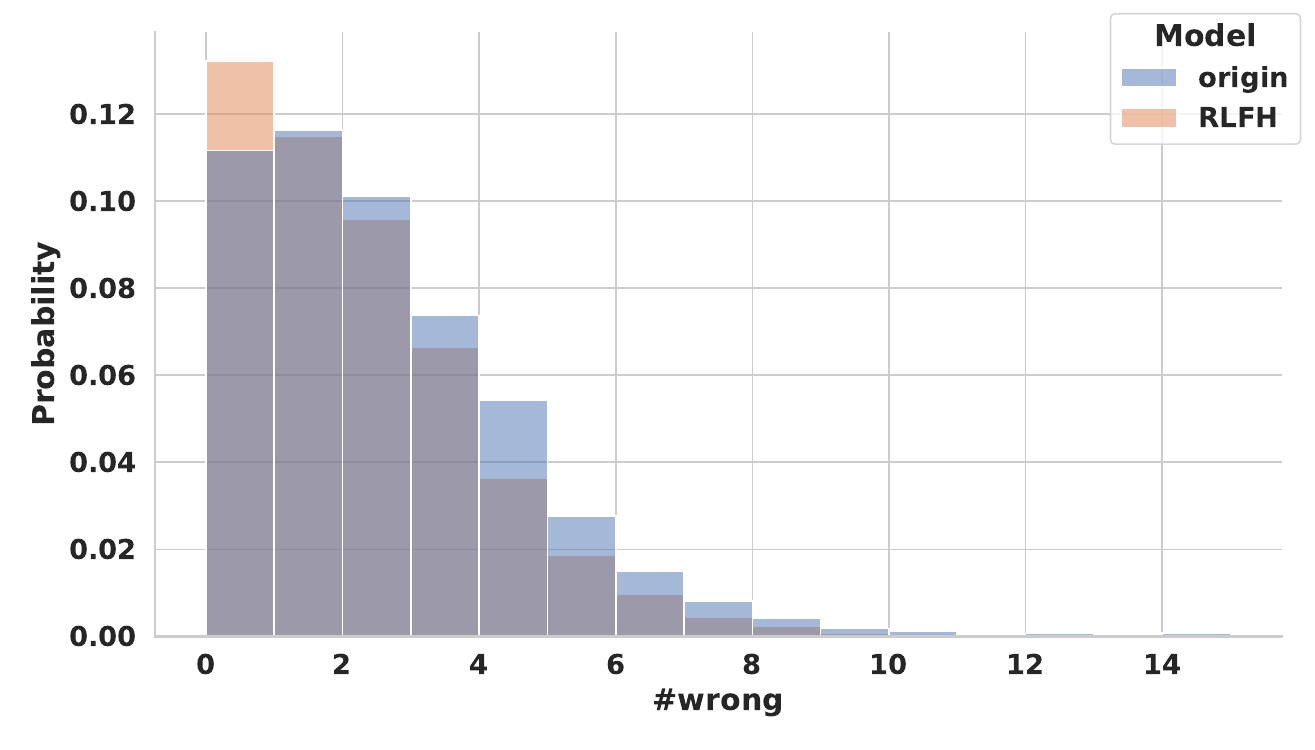}
    \label{fig:inc}
}
\caption{Distribution of statements per response across different truthfulness categories, comparing base Qwen2.5-7B-Instruct and its RLFH-tuned version. The distributions are normalized due to the filtering of rejected responses.}
\end{figure*}

\begin{figure}[t]
    \centering
    \includegraphics[width=\linewidth]{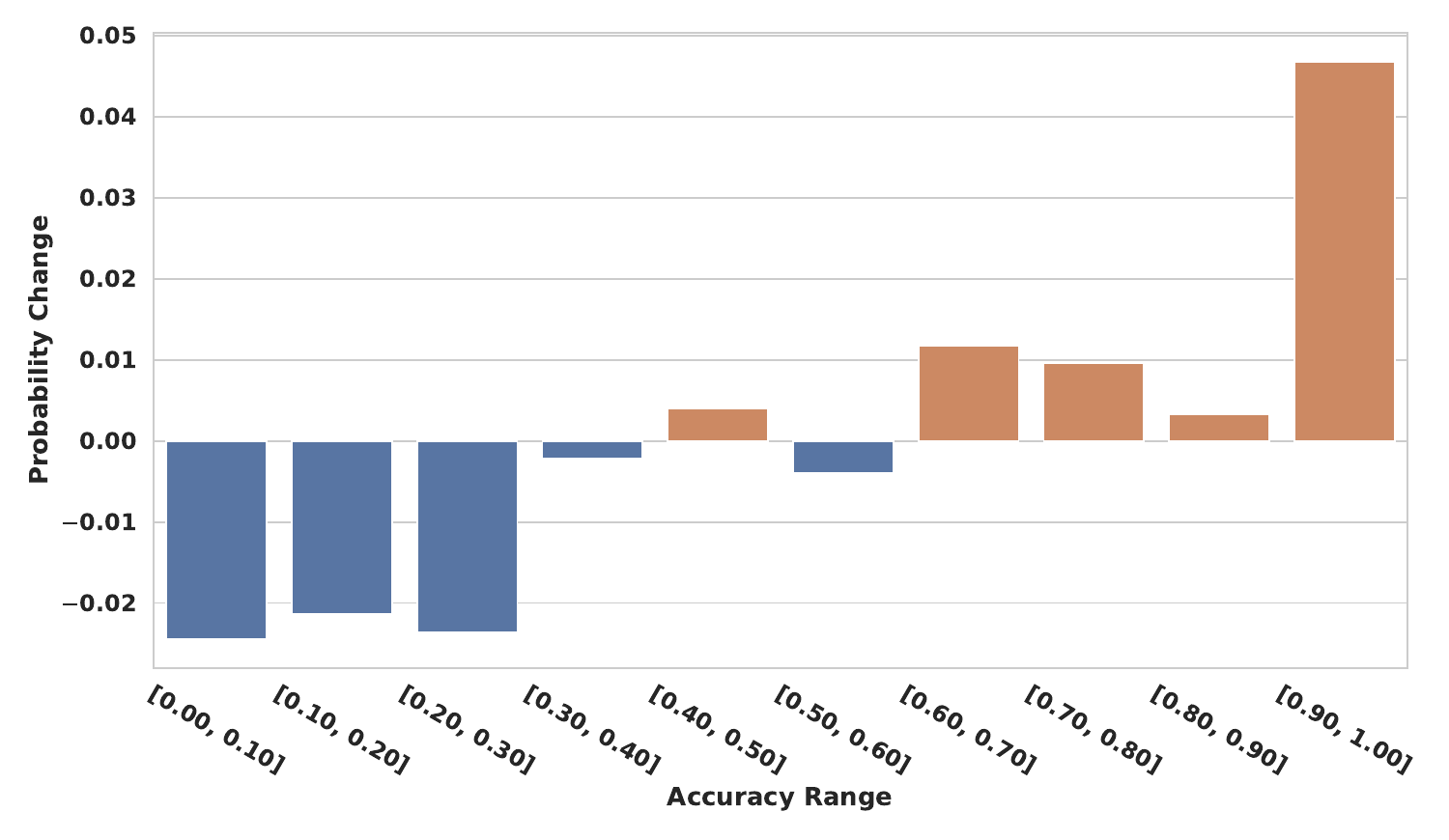}
    \caption{Response frequency distribution difference across statement accuracy for Qwen2.5-7B-Instruct, comparing the base model and RLFH-tuned model.}
    \label{fig:acc_diff}
\end{figure}

\paragraph{Our method significantly mitigates hallucination.}
The results show that our method achieves the highest FactScore across all datasets.
Given that FactScore is a well-established metric for assessing the factuality with external knowledge support, this consistent improvement in FactScore substantiates the effectiveness of our method.

\paragraph{The improvement is generalizable to out-of-distribution prompts.}
Notably, despite only being trained on the HotpotQA dataset, our algorithm demonstrated improved accuracy on two out-of-distribution datasets with different task settings.
This indicates that our method enables effective knowledge utilization as a generalizable capability across different tasks.
This generalizability suggests that RLFH improves the model's fundamental ability to assess and utilize its knowledge, rather than merely optimizing for specific dataset patterns.

\paragraph{The aligned model is generally more conservative but provides more accurate information within its capacity.} 
\label{sec:cons}
As shown in Table \ref{tab:main_result}, our trained model shows a decreased response ratio and a higher FactScore.
This observation aligns with expectations, as improving truthfulness often requires trading off helpfulness manifested by the reduced number of statements in the responses.
Figure \ref{fig:join} presents the joint distribution of accuracy and the number of statements shifts to the lower right direction, indicating that the model generates responses more conservatively while increasing the reliability of the information provided.

\begin{table*}[t]
\centering
\resizebox{\linewidth}{!}{%
\begin{tabular}{lccccccccccccccc}
\toprule
\multicolumn{1}{c}{\multirow{2}{*}{\textbf{Model}}} & \textbf{Avg.} & \multicolumn{4}{c}{\textbf{HotpotQA}} & & \multicolumn{4}{c}{\textbf{SQuADv2}} & & \multicolumn{4}{c}{\textbf{Biography}} \\
\cmidrule{3-6}\cmidrule{8-11}\cmidrule{13-16}
\multicolumn{1}{c}{} & \textbf{Score} & \textbf{\#Cor.} & \textbf{\#Inc.} & \textbf{\%Res.} & \textbf{Score} & & \textbf{\#Cor.} & \textbf{\#Inc.} & \textbf{\%Res.} & \textbf{Score} & & \textbf{\#Cor.} & \textbf{\#Inc.} & \textbf{\%Res.} & \textbf{Score} \\
\midrule
\multicolumn{16}{c}{\textit{RLFH Based On Qwen2.5-Instruct-7B Different Granularity Levels}} \\
\midrule
Qwen2.5-7B & 0.638 & 9.13 & 4.80 & 0.926 & 0.634 & & 21.09 & 4.82 & 0.974 & 0.813 & & 20.91 & 23.13 & 0.731 & 0.467 \\
Qwen2.5-7B$_\text{Response}$ & 0.651 & 8.09 & 4.31 & 0.910 & 0.639 & & 20.09 & 4.20 & 0.984 & 0.819 & & 20.24 & 19.96 & 0.654 & 0.493 \\
Qwen2.5-7B$_\text{Sentence}$ & 0.655 & 7.88 & 4.18 & 0.910 & 0.637 & & 18.87 & 3.86 & 0.974 & 0.821 & & 19.68 & 18.43 & 0.637 & 0.506 \\
Qwen2.5-7B$_\text{Statement}$ & \textbf{0.668} & 7.30 & 3.66 & 0.902 & \textbf{0.651} & & 17.29 & 3.55 & 0.963 & \textbf{0.830} & & 17.54 & 15.52 & 0.593 & \textbf{0.523} \\
\midrule
\multicolumn{16}{c}{\textit{RLFH Based On Llama3.1-8B-Instruct Different Granularity Levels}} \\
\midrule
Llama3.1-8B & 0.639 & 4.57 & 2.44 & 0.988 & 0.653 & & 22.75 & 6.02 & 0.984 & 0.777 & & 17.65 & 12.54 & 0.841 & 0.487 \\
Llama3.1-8B$_\text{Response}$ & 0.647 & 3.61 & 1.39 & 0.996 & 0.668 & & 17.71 & 4.61 & 0.990 & 0.758 & & 15.88 & 9.29 & 0.879 & 0.516 \\
Llama3.1-8B$_\text{Sentence}$ & 0.669 & 5.27 & 2.22 & 1.000 & 0.698 & & 21.63 & 5.29 & 0.990 & \textbf{0.789} & & 17.06 & 10.82 & 0.923 & 0.520 \\
Llama3.1-8B$_\text{Statement}$ & \textbf{0.686} & 6.23 & 2.10 & 1.000 & \textbf{ 0.714} & & 21.23 & 5.32 & 0.995 & 0.786 & & 17.30 & 10.98 & 0.791 & \textbf{0.558} \\
\bottomrule
\end{tabular}}
\caption{Results of baseline and RLFH-trained models using different granularity reward signals.}
\label{tab:dense}
\end{table*}

\begin{table*}[t]
\centering
\resizebox{\linewidth}{!}{%
\begin{tabular}{lccccccccccccccc}
\toprule
\multicolumn{1}{c}{\multirow{2}{*}{\textbf{Model}}} & \textbf{Avg.} & \multicolumn{4}{c}{\textbf{HotpotQA}} & & \multicolumn{4}{c}{\textbf{SQuADv2}} & & \multicolumn{4}{c}{\textbf{Biography}} \\
\cmidrule{3-6}\cmidrule{8-11}\cmidrule{13-16}
\multicolumn{1}{c}{} & \textbf{Score} & \textbf{\#Cor.} & \textbf{\#Inc.} & \textbf{\%Res.} & \textbf{Score} & & \textbf{\#Cor.} & \textbf{\#Inc.} & \textbf{\%Res.} & \textbf{Score} & & \textbf{\#Cor.} & \textbf{\#Inc.} & \textbf{\%Res.} & \textbf{Score} \\
\midrule
\multicolumn{16}{c}{\textit{RLFH Based on Qwen2.5-Instruct-7B with Different Judge Models }} \\
\midrule
Qwen2.5-7B & 0.638 & 9.13 & 4.80 & 0.926 & 0.634 & & 21.09 & 4.82 & 0.974 & 0.813 & & 20.91 & 23.13 & 0.731 & 0.467 \\
Qwen2.5-7B$_\text{DeepSeekV2-Lite}$ & 0.643 & 9.50 & 5.15 & 0.926 & 0.634 & & 21.64 & 5.08 & 0.979 & 0.802 & & 20.68 & 20.71 & 0.714 & 0.493 \\
Qwen2.5-7B$_\text{Llama3.1-8B}$ & 0.666 & 9.77 & 4.69 & 0.906 & 0.653 & & 21.93 & 4.51 & 0.979 & 0.814 & & 20.75 & 17.73 & 0.659 & \textbf{0.530} \\
Qwen2.5-7B$_\text{Qwen2.5-7B}$ & 0.668 & 8.31 & 3.99 & 0.906 & \textbf{0.655} & & 19.71 & 4.24 & 0.979 & 0.825 & & 19.09 & 16.92 & 0.604 & 0.523 \\
Qwen2.5-7B$_\text{On-Policy}$ & \textbf{0.668} & 7.30 & 3.66 & 0.902 & 0.651 & & 17.29 & 3.55 & 0.963 & \textbf{0.830} & & 17.54 & 15.52 & 0.593 & 0.523 \\
\midrule
\multicolumn{16}{c}{\textit{RLFH Based on Llama3.1-8B-Instruct with Different Judge Models }} \\
\midrule
Llama3.1-8B & 0.639 & 4.57 & 2.44 & 0.988 & 0.653 & & 22.75 & 6.02 & 0.984 & 0.777 & & 17.65 & 12.54 & 0.841 & 0.487 \\
Llama3.1-8B$_\text{DeepSeekV2-Lite}$ & 0.663 & 2.50 & 1.05 & 1.000 & 0.707 & & 10.51 & 2.53 & 1.000 & 0.762 & & 5.70 & 2.59 & 0.973 & 0.520 \\
Llama3.1-8B$_\text{Qwen2.5-7B}$ & 0.679 & 2.88 & 1.13 & 0.996 & 0.684 & & 10.37 & 2.47 & 0.990 & 0.782 & & 6.78 & 2.81 & 0.956 & 0.571 \\
Llama3.1-8B$_\text{Llama3.1-8B}$ & 0.675 & 3.35 & 2.09 & 0.996 & 0.677 & & 18.20 & 4.69 & 0.995 & 0.773 & & 15.06 & 8.94 & 0.830 & \textbf{0.575} \\
Llama3.1-8B$_\text{On-Policy}$ & \textbf{0.686} & 6.23 & 2.10 & 1.000 & \textbf{0.714} & & 21.23 & 5.32 & 0.995 & \textbf{0.786} & & 17.30 & 11.00 & 0.791 & 0.558 \\
\bottomrule
\end{tabular}}
\caption{Results of RLFH with different base models and judge models.}
\label{tab:judge}
\end{table*}

\subsection{Detailed Results}
\label{sec:detail}
We conducted a detailed analysis on 5,000 HotpotQA questions held out from training, evaluated using our annotation pipeline with Qwen2.5-72B-Instruct serving as the judge model.

\begin{figure}[t]
    \centering
    \includegraphics[width=\linewidth]{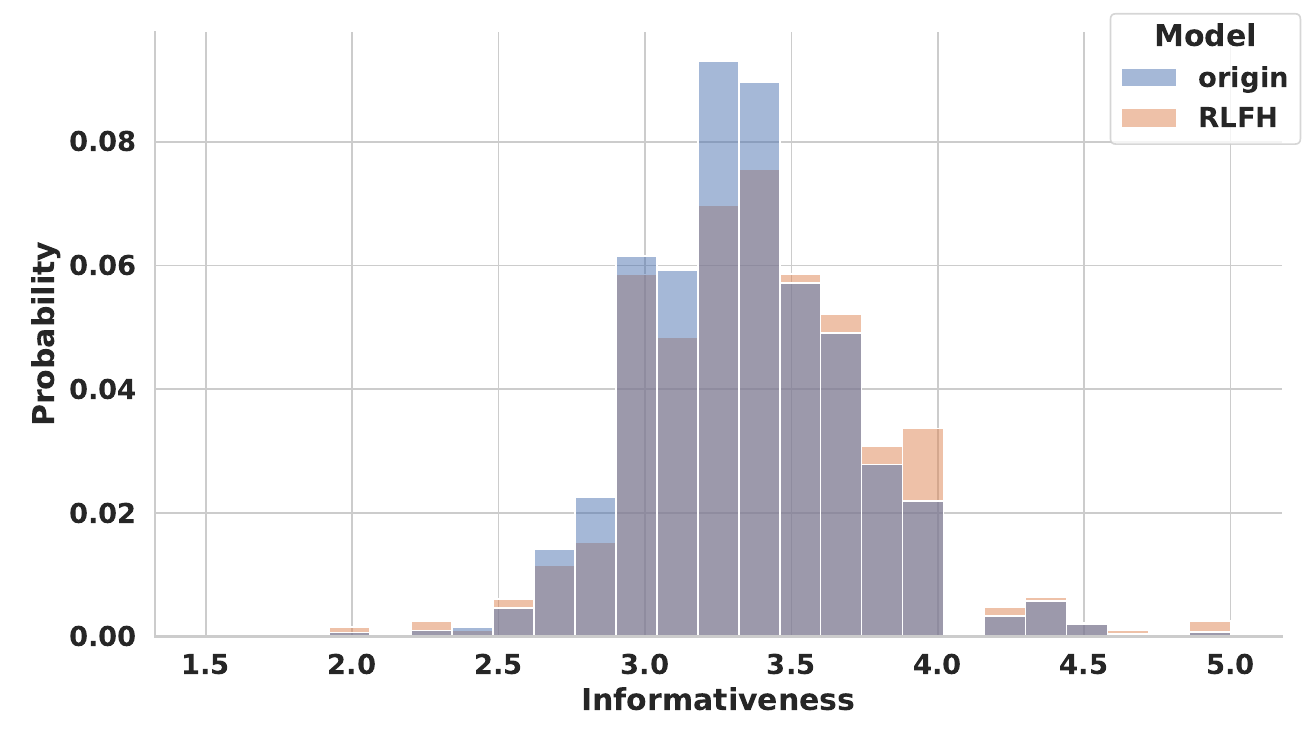}
    \caption{Frequency distribution of responses across different levels of average statement informativeness, comparing base and RLFH-tuned models.}
    \label{fig:info}
\end{figure}

\paragraph{Our method increases the ratio of high-accuracy responses.}
As shown in Figure \ref{fig:acc_diff}, our method decreases the proportion of low-accuracy responses and increases high-accuracy ones.
In particular, we observe a substantial increase in responses with accuracy exceeding 0.7, indicating that the model response is more reliable after training.
This improvement is attributable to RLFH's reward design, which generally penalizes responses with low accuracy while rewarding those with high accuracy.

\paragraph{Our algorithm suppresses errors and unverifiable content.}
The distribution shifts shown in Figures \ref{fig:vague} and \ref{fig:inc} indicate that RLFH effectively reduces both erroneous and unverifiable statements in model responses.
Meanwhile, as shown in Figure \ref{fig:inc}, the distribution of correct statements shows a trend towards generating a moderate number of statements.
This trend is expected, as increasing the number of statements raises the risk of errors, while fewer statements limit information coverage.

\paragraph{Our approach augments the average informativeness of statements in responses.}
As shown in Figure \ref{fig:info}, the response frequency distribution shifts towards higher informativeness, indicating that the model's responses generally provide more crucial information after training.
This proves that our model does not simply reduce information importance to minimize error probability.
Notably, we also observe a slight increase in the frequency of responses with low average informativeness.
This is reasonable since, the model tends to express more cautious responses or refuse to respond after tuning, which can be rated as less informative.

\subsection{Impact of Reward Granularity} 
\label{granu}
In this section, we conduct ablation experiments to investigate the impact of reward signal granularity.
Specifically, we evaluate paragraph-level, sentence-level, and statement-level reward.
The statement-level reward is our default setting described in previous sections.
For the sentence-level reward, feedback is incorporated at the end token of each sentence.
For the response-level reward, all feedback is aggregated into a single value for the entire response.
As shown in Table \ref{tab:dense}, statement-level rewards consistently achieve the highest FactScore, improving the average score from 0.638 to 0.668 for Qwen2.5-7B and from 0.639 to 0.686 for Llama3.1-8B.
This result underlines the importance of fine-grained feedback for developing more reliable models.

\subsection{Impact of Judge Model}
\label{sec:anno_model}
In this section, we explore the impact of different judge models in providing feedback signals.
Specifically, we compare two settings: on-policy setting where the policy model itself serves as the judge versus different fixed external judge models.
As shown in Table \ref{tab:judge}, for Qwen2.5-7B, the on-policy setting achieves the highest average score (0.668) along with fixed Qwen2.5-7B judge.
For Llama3.1-8B, the on-policy approach notably outperforms all fixed judge models, achieving the highest average score of 0.686.
The results validate the benefits of our on-policy setting, which not only achieves superior performance but also eliminates the need for an additional reward model in the training process.

\section{Conclusion}
In this work, we introduce an on-policy self-alignment approach that enables LLMs to explore their knowledge and self-correct hallucination behavior.
Our approach features a self-assessment framework where the policy serves as its own judge, automatically providing fine-grained feedback through atomic fact verification and generating token-level dense rewards for online reinforcement learning.
Comprehensive evaluations demonstrate that our approach effectively mitigates hallucination.
Our work represents a step towards more reliable and self-aware language models.

\section*{Limitations} \label{sec:limit}
Despite the promising results, our work has several limitations that warrant future investigation:
(1) Our work primarily addresses factual knowledge, while the broader challenge of generalized hallucination across diverse domains remains to be explored.
(2) Current evaluation benchmarks are limited in scope and may not fully capture the complex nature of hallucination, suggesting the need for more comprehensive evaluation frameworks.
(3) Although our self-alignment approach reduces the need for manual verification, the potential errors in automated fact-checking may still affect the optimal performance of the proposed method.

\section*{Broader Impacts} 
\label{sec:board}
Our research addresses a fundamental challenge in AI development by promoting truthful behavior in large language models through self-alignment, contributing to the development of more reliable AI systems.
By reducing hallucination in LLMs, our approach helps mitigate the spread of misinformation and enhances the models' utility in real-world applications, particularly as these models become increasingly integrated into society.
However, we acknowledge potential risks in relying solely on self-alignment mechanisms. 
The complete removal of human oversight in favor of AI self-verification could lead to inner alignment issues, where the model's learned behavior might deviate from intended objectives while appearing externally aligned.
A critical concern is that models might generate incorrect responses while simultaneously validating their own errors, creating a scenario where human verification becomes difficult or impossible. 
This self-reinforcing cycle could potentially lead to the propagation of misinformation and failure of alignment objectives.
We believe addressing these challenges requires deeper investigation into the interplay between self-alignment mechanisms and human oversight in ensuring reliable model behavior.

\section*{Acknowledgment}
We sincerely thank the reviewers for their insightful comments and valuable suggestions. This work was supported by Beijing Natural Science Foundation (L243006), Beijing Municipal Science and Technology Project (Nos. Z231100010323002), the Natural Science Foundation of China (No. 62306303, 62272439, 62476265) and the Basic Research Program of ISCAS (Grant No. ISCAS-JCZD-202303).


\bibliography{custom}

\appendix
\section{Prompt for AI Feedback}
\label{sec:prompt}
Following Section \ref{sec:feedback}, we design prompts for extraction, verification, and assessment tasks, as shown in Table \ref{tab:extract}, \ref{tab:verify}, and \ref{tab:assess}, respectively.

\begin{figure*}[ht]
\begin{tcolorbox}[colframe=cyan!40!black, title=\textbf{Statement Extraction Prompt}]
- Find every sentence containing object facts. \\
- Break sentences into atomic statements. \\
- Skip the sentences without statements. \\
- If there is no valid sentence, output "No statements". \\
- Do not output any explanation or other words. \\
- Strictly follow the output format shown in the example. \\
 \\
Here is an example: \\
\# Response \\
It is difficult to say which game has been released in more versions without more information, so I can only guess based on my training data. \\
Arthur's Magazine was likely started first. It was possibly founded in 1923 by Arthur K. Watson, a prominent publisher in the field of men's magazines. \\
First for Women, on the other hand, was not founded until 1989. It was created as a spin-off of Family Circle magazine, which was founded in 1957. \\
 \\
\# Statements \\
>> Sentence 1: Arthur's Magazine was likely started first. \\
* Arthur's Magazine was likely started first. \\
>> Sentence 2: It was possibly founded in 1923 by Arthur K. Watson, a prominent publisher in the field of men's magazines. \\
* Arthur's Magazine was possibly founded in 1923. \\
* Arthur's Magazine was founded by Arthur K. Watson. \\
* Arthur K. Watson is a prominent publisher in the field of men's magazines. \\
>> Sentence 3: First for Women, on the other hand, was not founded until 1989. \\
* First for Women was not founded until 1989. \\
>> Sentence 4: It was created as a spin-off of Family Circle magazine, which was founded in 1957. \\
* First for Women was created as a spin-off of Family Circle magazine. \\
* Family Circle magazine was founded in 1957. \\
 \\
And then comes your task: \\
\# Response \\
\{response\} \\
 \\
\# Statements
\end{tcolorbox}
\caption{Template for extracting statements from the model responses.}
\label{tab:extract}
\end{figure*}

\begin{figure*}[ht]
\begin{tcolorbox}[colframe=cyan!40!black, title=\textbf{Statement Verification Prompt}]
Choose from "Correct", "Vague" and "Wrong" for the verification of the statement. \\
- "Correct": The statement is supported by the materials. \\
- "Vague": Hard to determine the truthfulness of the statement based on the materials. \\
- "Wrong": The statement is negated by the materials. \\
Directly output the verification result without explanation. \\
Here is an example: \\
\\
\# Materials \\
- First for Women is a women's magazine published by Bauer Media Group in the USA. The magazine was started in 1989. It is based in Englewood Cliffs, New Jersey. In 2011 the circulation of the magazine was 1,310,696 copies. \\
- Arthur's Magazine (1844--1846) was an American literary periodical published in Philadelphia in the 19th century. Edited by T.S. Arthur, it featured work by Edgar A. Poe, J.H. Ingraham, Sarah Josepha Hale, Thomas G. Spear, and others. In May 1846 it was merged into "Godey's Lady's Book". \\
- The correct answer for the question "Which magazine was started first Arthur's Magazine or First for Women" may be "Arthur's Magazine". \\
\# Statement \\
Arthur's Magazine was likely started first. \\
\# Verification \\
Correct \\
\\
And then comes your task: \\
\# Materials \\
\{materials\} \\
\# Statement \\
\{statement\} \\
\# Verification
\end{tcolorbox}
\caption{Template for verifying statement based on external material.}
\label{tab:verify}
\end{figure*}

\begin{figure*}[ht]
\begin{tcolorbox}[colframe=cyan!40!black, title=\textbf{Statement Assessment Prompt}]
Evaluate the helpfulness of the statement: \\
- "5": The statement answer the question. \\
- "4": The statement provides crucial information. \\
- "3": The statement contains relevant facts. \\
- "2": The statement is about other supplementary facts. \\
- "1": The statement is useless or not relevant at all. \\
Directly output the evaluation result without explanation. \\
\\
Here is an example: \\
\# Question \\
Which magazine was started first Arthur's Magazine founded by Arthur K. Watson or First for Women? \\
\# Response \\
It is difficult to say which game has been released in more versions without more information, so I can only guess based on my training data. \\
Arthur's Magazine was likely started first. It was possibly founded in 1923 by Arthur K. Watson, a prominent publisher in the field of men's magazines. \\
First for Women, on the other hand, was not founded until 1989. It was created as a spin-off of Family Circle magazine, which was founded in 1957. \\
\# Statement \\
Arthur's Magazine was possibly founded in 1923. \\
\# Evaluation \\
4 \\
\\
And then comes your task: \\
\# Question \\
\{question\} \\
\# Response \\
\{response\} \\
\# Statement \\
\{statement\} \\
\# Evaluation
\end{tcolorbox}
\caption{Template for assessing statement importance based on original response.}
\label{tab:assess}
\end{figure*}

\section{Hyperparameters}
\label{sec:hyper}
We perform RLFH training with two different models: Qwen-2.5-7B-Instruct and Llama-3.1-8B-Instruct.
We adopt different hyperparameter settings for the two models according to their own characteristics.
For all experiments, we keep the same hyperparameter settings.

\begin{table}[ht]
\centering
\caption{Hyperparameter Settings for Different Models}
\resizebox{\columnwidth}{!}{
\begin{tabular}{lcc}
\hline
\textbf{Hyperparameter} & \textbf{Qwen-2.5-7B} & \textbf{Llama-3.1-8B} \\
\hline
Actor Learning Rate & 3e-7 & 5e-7 \\
Critic Learning Rate & 9e-6 & 9e-6 \\
KL Coefficient & 1e-2 & 5e-2 \\
Train Batch Size & 128 & 128 \\
Rollout Batch Size & 128 & 128 \\
Episode & 1 & 1 \\
Advantage Estimator & GAE & GAE \\
GAE $\lambda$ & 0.95 & 0.95 \\
Truthfulness Weight $\alpha$ & 1 & 1 \\
Informativeness Weight $\beta$ & 1.2 & 1.2 \\
Informativeness $\epsilon$ & -0.9 & -0.9 \\
Informativeness $\mu$ & 1.0 & 1.0 \\
\hline
\multicolumn{3}{l}{\textbf{Verification Map Function $f$}} \\
Correct & 0.45 & 0.2 \\
Hedged correct & 0.35 & 0.1 \\
Vague & -1.0 & -1.8 \\
Hedged wrong & -1.5 & -2.0 \\
Wrong & -1.7 & -2.2 \\
\hline
\multicolumn{3}{l}{\textbf{Informative Map Function $G$}} \\
5 & 1.25 & 1.2 \\
4 & 1.0 & 1.0 \\
3 & 0.75 & 0.8 \\
2 & 0.1 & 0.6 \\
1 & -0.2 & -0.1 \\
\hline
\end{tabular}
}
\end{table}

\section{Computation Resource}
\label{sec:compute}
While our PPO-based implementation requires additional resources compared to FACT's DPO approach \cite{tian2023finetuning}, including an additional value model and inference engines, the overall computational cost remains comparable for several reasons.
First, FACT requires multiple response samples per prompt, whereas our method needs only one.
Additionally, FACT relies on the FactScore pipeline \citep{min2023factscore} typically run by a larger annotation model (Qwen2.5-72B in our experiments), while our method utilizes the model being trained (Qwen2.5-7B or Llama3.1-8B in our experiments) for self-assessment, significantly reducing annotation costs.
In our experiments, each training run typically takes less than 1.5 hours using two 8-GPU nodes.
Alternatively, training on a single 8-GPU node requires approximately 3 hours per run.
Compared to traditional RLHF, our self-alignment approach eliminates the need for a separate reward model, resulting in reduced computational requirements.
Furthermore, while our implementation is based on PPO, more computationally efficient online reinforcement learning algorithms such as RLOO \citep{ahmadian2024basicsrevisitingreinforcestyle}, REINFORCE++ \citep{hu2025reinforcesimpleefficientapproach} or GRPO \citep{shao2024deepseekmathpushinglimitsmathematical} could be adopted.
These alternatives, which operate without a value model, would enable even more lightweight implementations of our approach.

\section{Factuality Metrics Validation}
\label{sec:metric_valid}
Following FactScore \cite{min2023factscore}, we validate the effectiveness of our hallucination evaluation. We conduct correlation analysis between Qwen2.5-72B-derived FactScore and human annotations across responses from three different models (InstructGPT, ChatGPT, and PerplexityAI).
The validation results, as shown in Table \ref{tab:main_result}, demonstrate strong alignment between our automated evaluation and human judgments.
Specifically, the Pearson correlation coefficients (COEF) between Qwen2.5-72B and human annotations are notably high: 0.923 for InstGPT, 0.909 for ChatGPT, and 0.737 for PPLAI.
Moreover, the error rates (ER) remain consistently low (0.041-0.098), which aligns with results from the original FactScore study.
These strong correlations and low error rates across different model responses validate the reliability of our evaluation approach.

\begin{table*}
\centering
\caption{Correlation analysis between human annotations and Qwen2.5-72B evaluations across different models.}
\label{tab:main_result}
\begin{tabular}{lccccccccc}
\toprule
\multirow{2}{*}{Evaluator} & \multicolumn{3}{c}{InstGPT} & \multicolumn{3}{c}{ChatGPT} & \multicolumn{3}{c}{PPLAI} \\
\cmidrule(lr){2-4} \cmidrule(lr){5-7} \cmidrule(lr){8-10}
& Score & COEF & ER & Score & COEF & ER & Score & COEF & ER \\
\midrule
Human & 0.428 & -- & -- & 0.583 & -- & -- & 0.804 & -- & -- \\
Qwen2.5-72B & 0.469 & 0.923 & 0.041 & 0.624 & 0.909 & 0.041 & 0.706 & 0.737 & 0.098 \\
\bottomrule
\end{tabular}
\end{table*}

\end{document}